\documentclass[letterpaper]{article} % DO NOT CHANGE THIS
\usepackage{aaai25}  % DO NOT CHANGE THIS
\usepackage{times}  % DO NOT CHANGE THIS
\usepackage{helvet}  % DO NOT CHANGE THIS
\usepackage{courier}  % DO NOT CHANGE THIS
\usepackage[hyphens]{url}  % DO NOT CHANGE THIS
\usepackage{graphicx} % DO NOT CHANGE THIS
\usepackage{subcaption}
\usepackage{amsmath}
\usepackage{natbib}  % DO NOT CHANGE THIS AND DO NOT ADD ANY OPTIONS TO IT
\usepackage{caption} % DO NOT CHANGE THIS AND DO NOT ADD ANY OPTIONS TO IT
\usepackage{algorithm}
\usepackage{algorithmic}

\usepackage{newfloat}
\usepackage{listings}
\DeclareCaptionStyle{ruled}{labelfont=normalfont,labelsep=colon,strut=off} % DO NOT CHANGE THIS
\floatstyle{ruled}
\newfloat{listing}{tb}{lst}{}
\floatname{listing}{Listing}
\title{From Unstable to Playable: Stabilizing Angry Birds Levels via Object Segmentation}
\author {
    Mahdi Farrokhimaleki,
    Parsa Rahmati,
    Richard Zhao
}
\affiliations {
    Department of Computer Science, University of Calgary\\
    Calgary, Alberta, Canada, T2N 1N4\\
    mahdi.farrokhimaleki@ucalgary.ca, parsa.rahmaty@ucalgary.ca, richard.zhao1@ucalgary.ca
}

\begin{document}

\maketitle

\begin{abstract}
Procedural Content Generation (PCG) techniques enable automatic creation of diverse and complex environments. While PCG facilitates more efficient content creation, ensuring consistently high-quality, industry-standard content remains a significant challenge. In this research, we propose a method to identify and repair unstable levels generated by existing PCG models. We use Angry Birds as a case study, demonstrating our method on game levels produced by established PCG approaches. Our method leverages object segmentation and visual analysis of level images to detect structural gaps and perform targeted repairs. We evaluate multiple object segmentation models and select the most effective one as the basis for our repair pipeline. Experimental results show that our method improves the stability and playability of AI-generated levels. Although our evaluation is specific to Angry Birds, our image-based approach is designed to be applicable to a wide range of 2D games with similar level structures.
\end{abstract}


\section{Introduction}

Procedural Content Generation (PCG) is the algorithmic creation of game content, such as levels, maps, and characters, with limited or indirect user input. PCG has become an essential tool in game development, enabling the automatic creation of diverse and complex environments \cite{summerville2018procedural, maleki2024procedural}. While PCG methods can generate vast amounts of content efficiently, ensuring playability and structural stability remains a critical challenge, particularly in physics-based games such as Angry Birds. In these games, levels consist of interconnected structures, where even minor instability can lead to unintended collapses, affecting the playability of levels.

Previous research in level generation has explored various approaches, including search-based methods, machine learning models, and user-adaptive techniques \cite{nygren2011user}. While traditional PCG methods primarily focus on generating levels from predefined rules, Generative adversarial networks (GANs) have recently emerged as a powerful alternative for learning structural patterns from human-designed levels \cite{abraham2023utilizing}. However, GAN-generated levels often lack explicit stability constraints, leading to floating blocks, misaligned structures, or physically unplayable configurations.

After reviewing the generated levels from Abraham and Stephenson (\citeyear{abraham2023utilizing})'s work, we observed that some unstable levels contain noticeable gaps that could be corrected using a model trained for gap detection. To address this, we propose a level repair pipeline that integrates object segmentation with automated level correction. Our method utilizes YOLOv8-Seg, a real-time object segmentation model, to identify gaps and unstable structures in procedurally-generated Angry Birds levels.

We define a stable level as one that does not exhibit unintended motion or block destruction when loaded into the game engine. Once gaps are identified, our method repairs them by inserting appropriate structural elements, thereby improving both the stability and playability of procedurally generated content.

For our study, we select GAN-generated Angry Birds levels and apply our repair method. Our approach introduces a novel segmentation-driven correction process that promotes structural coherence and reduces reliance on manual intervention. We evaluate the repaired levels using the stability metrics introduced in prior work, comparing the playability of original versus repaired levels through quantitative stability assessments.

%\subsection{Research Contributions}

This paper presents the following key \textbf{research contributions}:

\begin{enumerate}
    \item A segmentation-based repair pipeline for correcting unstable procedurally-generated Angry Birds levels.
    \item A real-time object segmentation model using YOLOv8, trained to identify gaps in Angry Birds structures.
    \item A comparative evaluation of level stability, demonstrating improvements in playability after repairs.
\end{enumerate}

The findings demonstrate that, despite achieving success in fixing only a portion of unplayable levels, segmentation-based level repair appreciably increases the aggregate count of playable levels when deployed within a large-scale generate-and-test pipeline.

\section{Background and Related Works}

\subsection{Level Repairing}
PCG is widely used in game development to create dynamic and diverse content, but ensuring high-quality, playable levels remains a challenge. Automated level repairing techniques have been introduced to detect and fix unstable or unplayable elements in AI-generated levels. This process involves identifying structural flaws, applying fixes, and re-evaluating stability to ensure functional and engaging gameplay.

Several approaches have been proposed for repairing procedurally generated levels, focusing on different aspects of stability, fairness, and playability. Rule-based correction with predefined rules has been used to remove floating blocks, prevent gaps, and ensure reachable objectives \cite{graves2016procedural}. Machine learning-based methods, such as neural networks, learn the probability distribution of tiles based on their surroundings \cite{shu2020novel}.

Despite these methods, unlike tile-based platformers where level geometry is discrete and predictable, level repairing in physics-based games remains particularly challenging due to the complex interactions between game objects. Angry Birds, being a physics-based game, introduces additional complexity in level repair. Structures in the game are composed of blocks with different materials (wood, ice, stone), and their interactions determine stability and destructibility. GAN-generated levels for Angry Birds, while effective in creating new structures, often suffer from instability problems, resulting in a low count of stable levels compared to unstable ones \cite{abraham2023utilizing}.

Our study aims to address this problem using a machine-learning-based approach, integrating YOLO-based object segmentation to identify gaps in unstable Angry Birds levels and applying automated repairs to enhance their stability and playability. By focusing on stabilizing unstable levels rather than generating stable levels from scratch, we aim to address one of the key barriers that limits the adoption of academic content generators. While many works suggest potential industry relevance, few are used in practice due to quality concerns. Improving stability through repair helps narrow this gap by enhancing the usability of such generators not only for potential industry use, but also in educational contexts and co-creative systems, where reliability and playability are equally important.

\subsection{Angry Birds Level Generation }
Angry Birds is a 2D physics-based game where players launch birds at block-based structures to eliminate pigs. The game mechanics rely on destructible environments, where structures are usually composed of wood, ice, or stone blocks, each having different resistance and attributes. Players must strategically aim their limited number of birds to cause structural collapse and maximize damage to eliminate all pigs. Because the original Angry Birds game is not open source, most academic research and AI-driven level generation studies utilize a Unity-based clone called Science Birds \cite{ferreira2014search}. Science Birds replicates the mechanics of the original game while allowing researchers to generate, modify, and analyze levels programmatically. Each level in Science Birds is stored in an XML format, describing the position, type, and orientation of every game object.

Table 1 presents the standard block types used in Angry Birds levels, categorized by shape and material. While the game also includes irregular block types, such as triangular and circular blocks, these are typically excluded from automated level generation due to complex stability constraints. Most PCG-based approaches for Angry Birds generate structures using only rectangular blocks with limited block orientations (0° or 90° angles) to minimize instability.

\begin{table}[ht]
    \centering
    \renewcommand{\arraystretch}{1.6} % less vertical spacing
    \setlength{\tabcolsep}{9pt} % less horizontal spacing
    \resizebox{1\linewidth}{!}{ % scale to 100% of text width
        \begin{tabular}{|c|c|c|c|}
            \hline
            \textbf{ID} & \textbf{Shape} & \textbf{Name} & \textbf{Dimensions} \\
            \hline
            1 & \makebox[1.2cm][c]{\raisebox{-0.5\height}{\includegraphics[width=0.6cm,height=0.6cm]{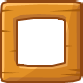}}} & SquareHole & (0.85, 0.85) \\ \hline
            2 & \includegraphics[width=1.2cm,height=0.15cm]{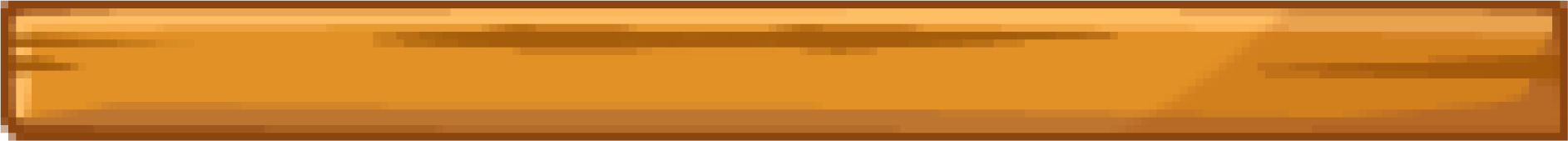} & RectBig & (2.06, 0.22) \\ \hline
            3 & \includegraphics[width=0.9cm,height=0.15cm]{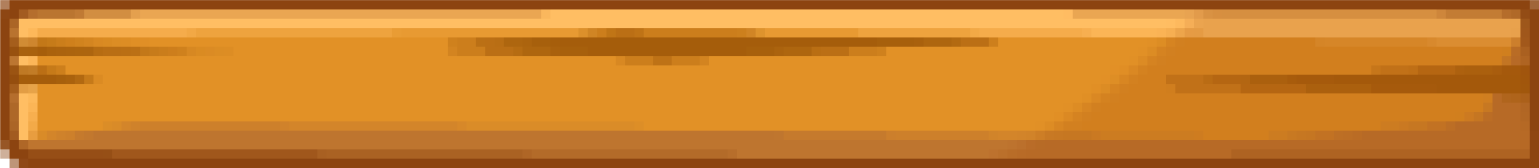} & RectMedium & (1.68, 0.22) \\ \hline
            4 & \includegraphics[width=0.6cm,height=0.15cm]{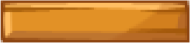} & RectSmall & (0.85, 0.2) \\ \hline
            5 & \includegraphics[width=0.6cm,height=0.3cm]{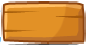} & RectFat & (0.85, 0.43) \\ \hline
            6 & \includegraphics[width=0.3cm,height=0.15cm]{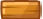} & RectTiny & (0.42, 0.22) \\ \hline
            7 & \includegraphics[width=0.15cm,height=0.15cm]{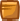} & SquareTiny & (0.22, 0.22) \\ \hline
            8 & \includegraphics[width=0.3cm,height=0.3cm]{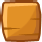} & SquareSmall & (0.43, 0.43) \\
            \hline
        \end{tabular}
    }
    \caption{Block types that are available in Science Birds.}
    \label{tab:my_table}
\end{table}

\subsubsection{Existing Approaches}

Over the past decade, researchers have explored multiple approaches for automated level generation in Angry Birds. Many of these methods have been showcased in the AIBirds Level Generation Competition \cite{stephenson20182017}. The most common techniques include: genetic algorithms (GA) that use evolutionary principles to iteratively improve level design \cite{ferreira2014search}, search-based approaches that generate levels by optimizing specific structural constraints \cite{stephenson2017generating,stephenson2016proceduralAIIDE}, Monte Carlo Tree Search (MCTS) that simulates multiple playthroughs to identify optimal level configurations \cite{graves2016procedural}, and variational autoencoders (VAE) that use deep learning to generate levels from latent representations \cite{tanabe2021level}. Some studies have focused on specific objectives, such as: generating levels that resemble text, quotes, or formulas \cite{jiang2017procedural}, creating deceptive structures that mislead players \cite{gamage2021deceptive}, and designing Rube Goldberg-style contraptions for dynamic interactions \cite{abdullah2019angry}.

Despite these advancements, ensuring stability remains a challenge, as AI-generated levels can often contain floating blocks, structurally unsound elements, or unintended difficulty spikes. This limitation highlights the need for post-processing techniques (such as the repair approach proposed in our study) to refine AI-generated levels and improve their playability.
Additionally, while some level generation approaches, such as the Rube Goldberg generator and the Deceptive Generator, report very high stability rates, these systems rely heavily on constrained templates and looser definitions of stability. Such methods produce simpler, more predictable structures where instability is far less likely to occur. In contrast, GAN-generated levels are more diverse and complex, which naturally leads to a higher rate of unstable structures. Our focus on GAN outputs was intentional: they represent a more challenging but also more promising direction for producing varied and creative levels. Importantly, instability is not unique to GAN-based methods. any approach that generates more intricate structures without strict constraints is prone to similar issues.

\subsection{Object Segmentation}
Object segmentation is a computer vision technique that involves identifying and delineating objects within an image at the pixel level \cite{long2015fully}. Unlike object detection, which provides bounding boxes around objects, segmentation classifies each pixel, enabling precise localization of structural components.

Deep learning-based segmentation models, particularly convolutional neural networks (CNNs), have demonstrated state-of-the-art performance in this field. Among them, YOLO (You Only Look Once) has emerged as one of the most efficient architectures for real-time segmentation and detection \cite{long2015fully}. YOLO, developed by Joseph Redmon et al. (\citeyear{redmon2016you}), was the first real-time, end-to-end object detection approach. The name ``You Only Look Once" highlights its ability to perform object detection in a single network pass, unlike earlier methods that required multiple evaluations per image. Previous techniques, such as sliding window-based approaches, involved applying a classifier hundreds or thousands of times across an image, making them computationally expensive. More advanced models, like two-stage detectors, first generated region proposals before running a classifier on each proposal. YOLO, in contrast, streamlined the detection process by using a regression-based approach to directly predict object locations and class probabilities in one step, unlike Fast R-CNN \cite{ren2016faster}, which relied on separate outputs for classification and bounding box regression. This design made YOLO significantly faster and more efficient than prior detection frameworks \cite{terven2023comprehensive}.

Among different versions of YOLO, YOLOv8 offers significant improvements in segmentation accuracy, computational efficiency, and model robustness \cite{huang2023research}.
The advancements convinced us to use this model in our research for our gap detection, where structures contain varied shapes, materials, and complex spatial arrangements.

We chose segmentation because it allows classification at the pixel level and enables detection of structures even when blocks overlap or are partially occluded. This capability was crucial for identifying meaningful gaps. A similar strategy was used in for Mario level repair \cite{shu2020novel}, demonstrating its relevance to games.

\section{Methodology}

This section outlines our pipeline for the automated repair of AI generated levels, using levels from Angry Birds generated using the Angry Birds GAN model \cite{abraham2023utilizing}. The pipeline begins by analyzing an image of a generated level to assess its structural stability. Unstable levels are then automatically passed to our trained repairer model for correction.

\subsection{Model Architecture}
For the gap detection task, we selected three distinct, state-of-the-art model architectures for comparison: U-Net, SegFormer, and YOLOv8m-Seg, the medium variant of the YOLOv8 segmentation model. This selection was made to evaluate a representative cross-section of modern segmentation paradigms. U-Net was chosen as a highly effective and well-established benchmark for convolutional-based segmentation. SegFormer represents the more recent transformer-based approach, which has achieved strong performance in various computer vision tasks. Finally, YOLOv8m-Seg was included for its foundation in real-time object detection, offering a lightweight and computationally efficient alternative. By comparing these three, we aimed to identify the most suitable architecture balancing accuracy and performance for our level repair pipeline.

SegFormer is a transformer-based architecture introduced by Xie et al. (\citeyear{xie2021segformer}). It combines the advantages of hierarchical transformers and lightweight multilayer perceptrons (MLPs) for efficient and accurate semantic segmentation. Unlike traditional convolutional models, SegFormer does not rely on positional encoding and uses a Mix-FFN and overlapping patch merging to capture both local and global context effectively. 

U-Net, proposed by Ronneberger et al. (\citeyear{ronneberger2015unet}), is a widely used convolutional network in medical and binary segmentation tasks. It consists of a symmetric encoder-decoder structure with skip connections that help preserve spatial information lost during downsampling. The encoder extracts features at multiple scales, while the decoder gradually reconstructs the segmentation mask, incorporating features from corresponding encoder levels.

YOLOv8-Seg is a recent advancement in the YOLO series developed by Ultralytics (\citeyear{ultralytics2023yolov8}). YOLOv8-Seg extends real-time object detection capabilities to segmentation by combining bounding box detection with precise mask prediction. It employs a lightweight and fast architecture that is suitable for deployment scenarios where inference speed is crucial. We used the YOLOv8m-Seg model with its default configuration and fine-tuned it for our binary segmentation task.

Each model was trained and evaluated on the same dataset and under similar conditions to ensure a fair comparison in terms of performance metrics and computational efficiency. Following the comparative analysis detailed in the Results section, YOLOv8m-Seg was selected as the definitive model for our repair pipeline. It provided the best combination of high segmentation accuracy and computational efficiency, making it the most practical choice for integration into our automated repair system.

\subsection{Dataset}
For training, \citeauthor{abraham2023utilizing} used the open-source level generator Iratus Aves \cite{stephenson2017generating} to produce a dataset of 4,931 XML level descriptions, each representing a unique structure, including pigs and a variety of blocks. We used the same dataset as our training data.

Given our focus on teaching the model to identify gaps that destabilize structures, we filtered these levels to select only those that were stable. This was essential because our goal was to train a model on de-stabilized versions of otherwise stable levels, enabling supervised learning with known ground-truth gaps that could be filled to restore stability.

We evaluated level stability using three different metrics derived from physical simulation. To ensure higher quality and consistency in training, we selected the most stringent stability metric, as it yielded the most robust set of stable levels. This filtered set became our stable dataset, and we reduced the level count to 1887 XML levels.

To generate training samples with structural gaps, we artificially introduced instability by removing one to four random blocks from the XML file of each selected level. We then simulated these modified levels, keeping only those where the original level was stable but the modified version became unstable after block removal. This filtering process resulted in 1,547 levels, which constituted our final dataset for training purposes.

For model training, each level's XML file was processed using the established encoding pipeline. The level was converted into a multi-layer grid representation, which was then flattened into a single 2D image. To focus the model on structure rather than material composition, this image was converted into a binary format (where 1 represents any object and 0 represents empty space). This binary image of the unstable structure served as the input for our model, while the pixel-perfect location of the removed block(s) served as the target segmentation mask.

This procedure ensured that each training sample corresponded to specific and known destabilizing gaps, aligning with our goal of using supervised learning to detect and suggest gap-filling interventions.
For our experiments, we utilized the dataset of 8,000 XML Angry Birds levels as presented and analyzed in the work of Abraham and Stephenson. This dataset was selected for two primary reasons. First, the authors noted significant instability issues within their generated levels, which provided an ideal test case for our repair pipeline. Second, using their established dataset allows for a more direct comparison of the improvements offered by our method. We acknowledge that these datasets design may not capture all possible structural failure modes. These limitations are discussed further in the Limitations section.

\begin{figure*}[htb]
    \centering
    \subfloat[\centering Repair Pipeline]{
        \includegraphics[width=1\textwidth]{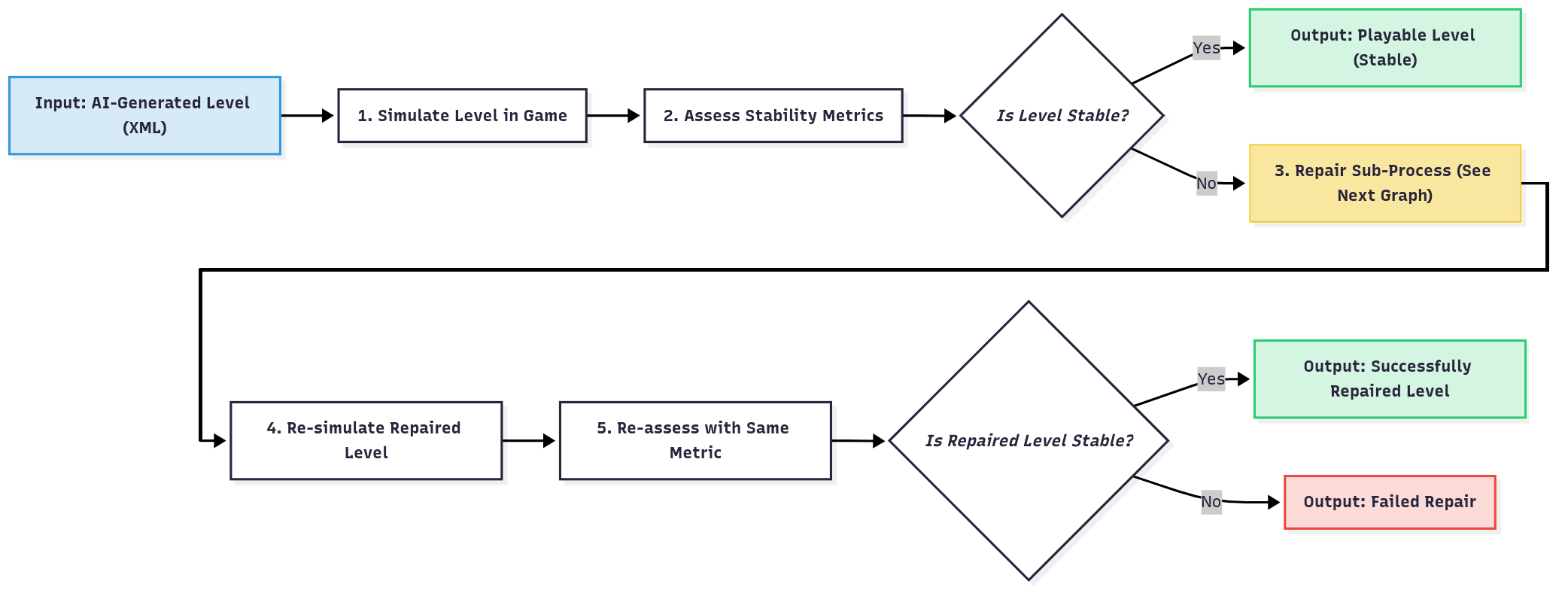}
    }\\[1ex]
    \subfloat[\centering Repair Sub-Process]{
        \includegraphics[width=1\textwidth]{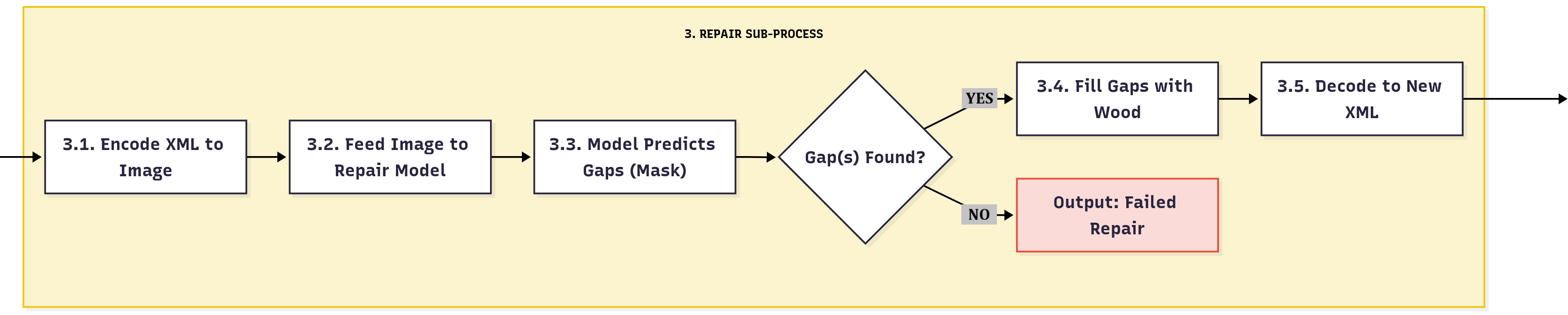}
    }
    \caption{The automated level repair pipeline. The top diagram (a) illustrates the high-level workflow, where an AI-generated level is first simulated and evaluated for stability. If the level is deemed unstable, it enters the repair stage. After the repair is applied, the level is re-evaluated to confirm its stability. The bottom diagram (b) provides a detailed breakdown of the ``Repair Sub-Process" block, showing the five steps from encoding the level into an image, using the model to predict gaps, filling the gaps with wood, and decoding the result back into a new XML file.}
    \label{fig:pipeline}
\end{figure*}

\subsection{Training Details}

\subsubsection{Training Environment}
 Model training was performed on a Google Colab instance, which provided access to a NVIDIA T4 GPU and 16 GB of RAM. This cloud-based environment was used for its powerful computational capabilities, suitable for training deep learning models.

\subsubsection{Training Hyper-parameters and Configuration}
 The evaluation, which necessitated the execution of the Science Birds simulator, was performed on a local Windows-based machine with the following specifications: CPU: Intel Core™ i5-10400F, GPU: NVIDIA GeForce 1660 Super, RAM: 16 GB.

Across all models, a consistent set of training parameters was used to ensure fair comparison. The dataset was partitioned into training (80\%) and validation (20\%) sets. The models were trained for a maximum of 100 epochs with a  batch size of 8 using the AdamW optimizer, a variant of the Adam optimizer that enhances weight decay regularization. Input images were resized to 128x128 pixels. An early stopping strategy was employed with a patience of 10, halting training if no improvement was observed after 10 consecutive epochs to mitigate overfitting. Finally, model weights were saved periodically every 5 epochs. Upon completion of training, the models were compared based on their segmentation performance on the validation dataset. To measure the precise quality of the generated masks against the ground truth, we calculated several pixel-level metrics such as Precision, Recall, and F1-Score for each model.

\begin{figure*}[htb]
    \centering
    \makebox[\textwidth][c]{
        \begin{subfigure}{0.4\textwidth}
            \centering
            \includegraphics[width=\textwidth]{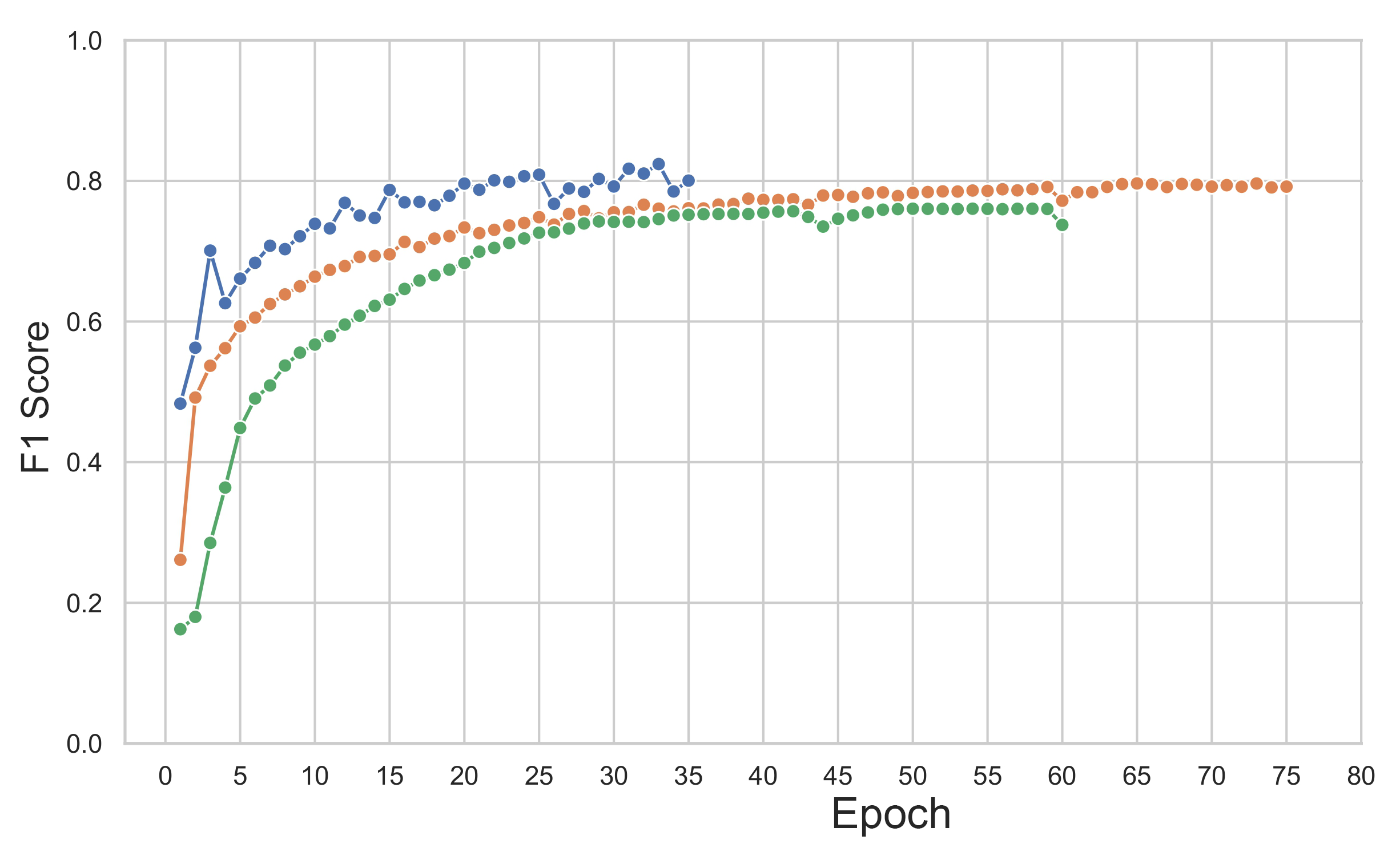}
            \caption{F1 Score over Epochs}
        \end{subfigure}%
        \hfill
        \begin{subfigure}{0.4\textwidth}
            \centering
            \includegraphics[width=\textwidth]{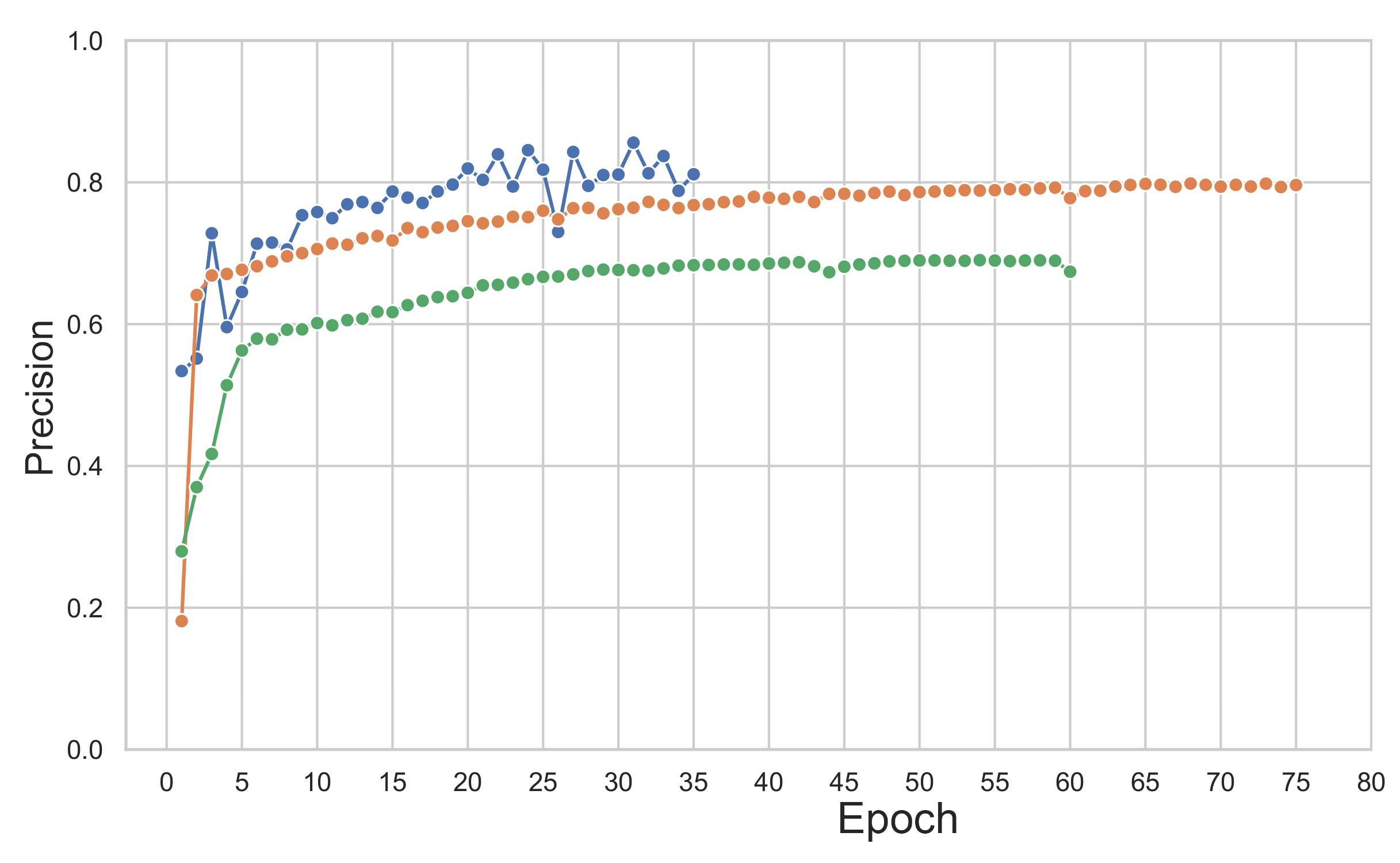}
            \caption{Precision over Epochs}
        \end{subfigure}
    }
    
    \makebox[\textwidth][c]{
        \begin{subfigure}{0.4\textwidth}
            \centering
            \includegraphics[width=\textwidth]{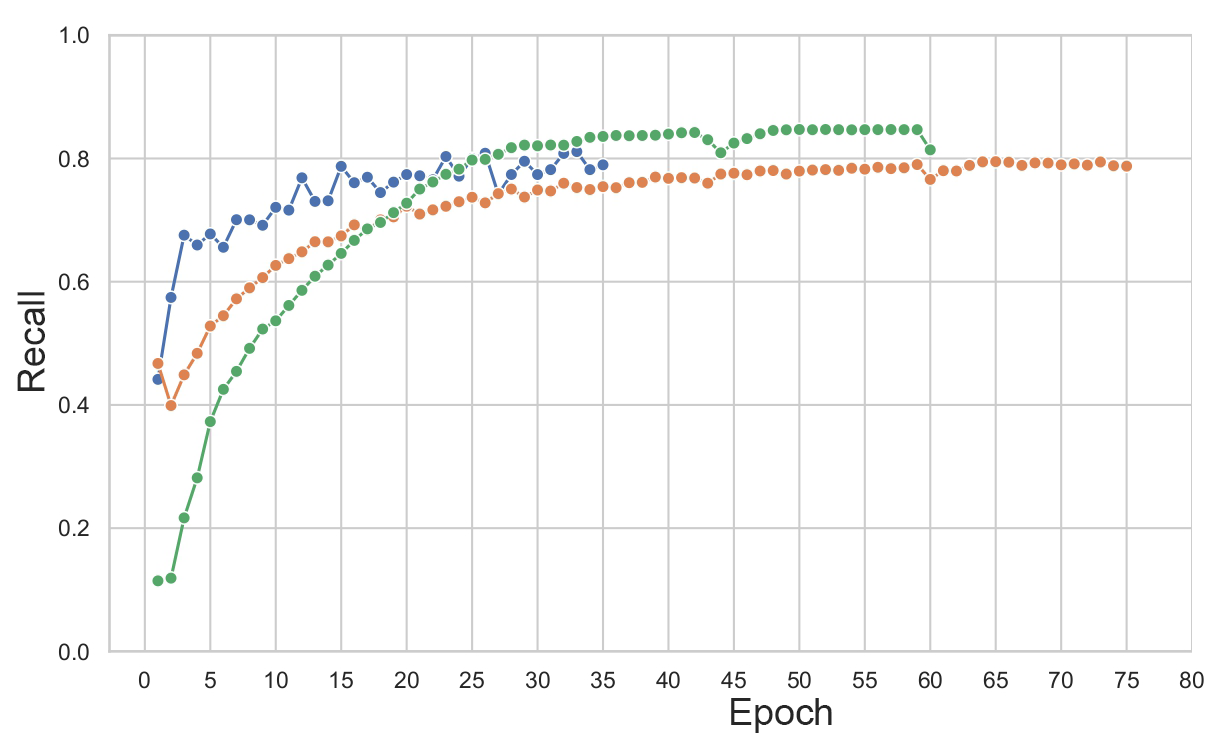}
            \caption{Recall over Epochs}
        \end{subfigure}%
        \hfill
        \begin{subfigure}{0.4\textwidth}
            \centering
            \includegraphics[width=\textwidth]{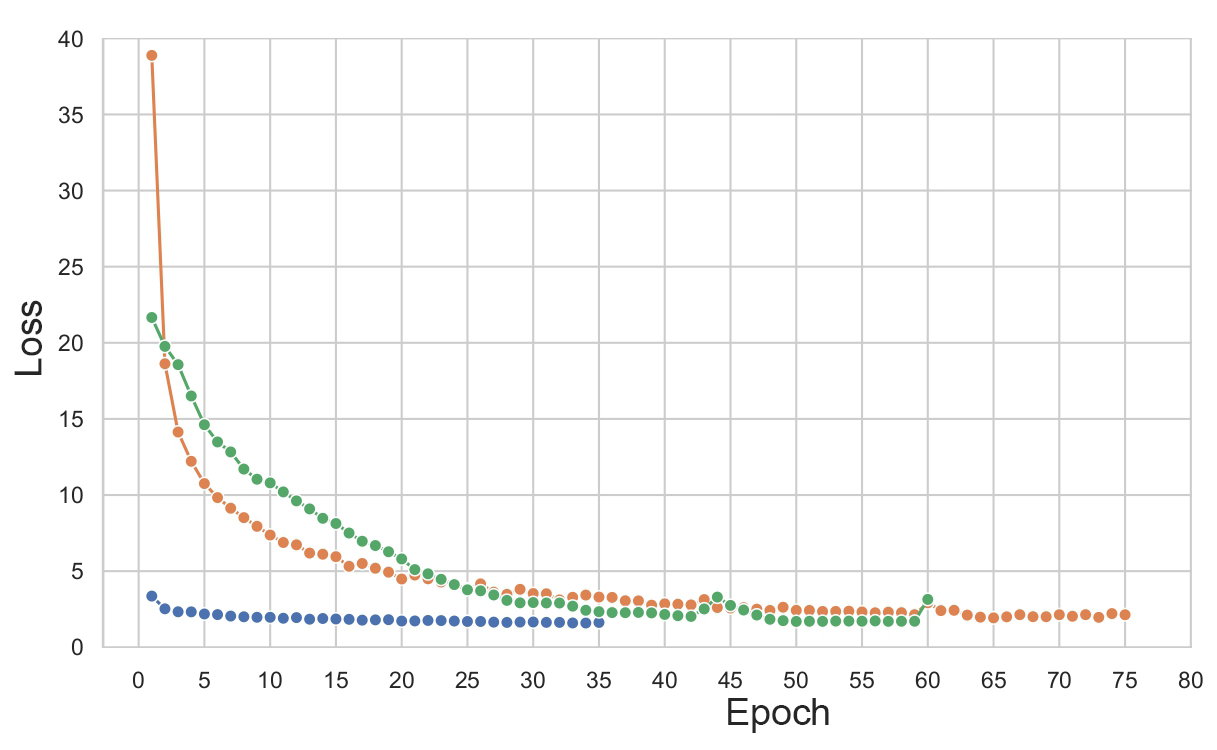}
            \caption{Loss over Epochs}
        \end{subfigure}
    }

    \caption{Training Progress of YOLOv8m (blue), SegFormer (orange), and UNet (green) models over epochs.}
    \label{fig:training-metrics}
\end{figure*}

\subsection{Repair Process}
Our automated repair pipeline leverages the trained model to correct instabilities in AI-generated levels. This process involves a three-stage evaluation, repair, and re-evaluation workflow. Figure \ref{fig:pipeline} illustrates our complete workflow.

The repair process starts by simulating each generated level within the game engine. During this simulation, the level's stability is evaluated against one of three predefined metrics: Block Velocity, Block Destruction, or Block Damage. If a level is identified as unstable according to these criteria, the repair stage is initiated. Here, a binary-encoded image of the level, which distinguishes structure from background, is provided as input to our trained model. The model produces a segmentation mask identifying the precise locations of structural gaps. These repairs are applied directly to a 2D image representation of the level by filling the pixel positions of the predicted gaps, which are then converted by the decoder into the most closely matching blocks..

For all repairs, wood is selected as the default material. This choice is based on its balanced physical properties compared to ice and stone. Our preliminary experiments reveal that material choice had a negligible effect on stability outcomes; the primary factors are the placement of blocks and their resulting physical interactions. Wood provides adequate structural integrity without the excessive weight of stone, which could introduce new instabilities, or the low friction and durability of ice.

In the final stage, the repaired level representation is decoded back into the standard XML format. This new level is then subjected to a final re-evaluation inside the game using the same metric that initially flagged it as unstable. This step serves to quantitatively verify whether the repair was successful and if the level has been stabilized.

\section{Evaluation and Results}

This section presents a two-part empirical evaluation of our proposed level repair pipeline. First, we conduct a comparative analysis of the three selected segmentation models, SegFormer, U-Net, and YOLOv8m-Seg, to identify the most effective architecture for the gap-detection task. Second, using the best-performing model, we evaluate the efficacy of the complete repair process when applied to the dataset of unstable levels generated by the GAN from Abraham and Stephenson (\citeyear{abraham2023utilizing}). 

The primary evaluation of our work measures the efficacy of the final repair model on the dataset of AI-generated levels. The success of the repair pipeline is quantified by the number of previously unstable levels that are rendered stable, as measured by our stability metrics explained in the next section.

To further understand the model's impact, we also analyze the cases where the repair was not successful. For the levels that remained unstable, we calculate the average block damage and the average number of destroyed blocks. These figures are compared against the baseline averages of the original, unrepaired unstable levels from the methods of Abraham and Stephenson, allowing us to assess if our model had a partially mitigating effect even in failed cases.

\subsection{Evaluation Metrics}
To conduct a thorough evaluation, we employ two distinct categories of metrics: one to assess the pixel-level accuracy of the segmentation models and another to measure the physical stability of the game levels after repair.

\subsubsection{Model Performance Metrics}
To compare the performance of the segmentation architectures, we use standard metrics that evaluate the quality of the predicted segmentation masks against the ground-truth data.

Precision, Recall, and F1-Score on pixel-level metrics assess the accuracy of the segmentation. Precision measures the proportion of correctly predicted gap pixels among all pixels predicted as a gap. Recall measures the proportion of correctly predicted gap pixels among all actual gap pixels. The F1-Score is the harmonic mean of Precision and Recall.

After selecting YOLOv8m-Seg as our final model, we utilize additional metrics, common in object detection, for a more precise evaluation of its ability to locate the exact position and shape of a gap. Intersection over Union (IoU), also known as the Jaccard index, measures the overlap between the predicted segmentation mask and the ground-truth mask, and it is calculated as the area of their intersection divided by the area of their union. For segmentation tasks, mean Average Precision (mAP) provides a comprehensive measure of a model's accuracy across various IoU thresholds. mAP50 calculates the average precision at a fixed IoU threshold of 0.50, while mAP50-95 calculates the average precision over a range of IoU thresholds from 0.50 to 0.95 in steps of 0.05, offering a more accurate evaluation of the mask prediction quality.

\subsubsection{Level Stability Metrics}
To quantify the success of the repair process, we assess the stability of levels within the game engine. We adopt the two metrics from Abraham and Stephenson's work and introduce a third to capture more nuanced outcomes.

The “Block Velocity” measure determines that a structure is stable if all blocks are stationary when the level is loaded. The “Block Destruction” measure determines that a structure is stable if no blocks are destroyed after the level is loaded. Blocks in Angry Birds will typically be destroyed if they fall from a sufficient height or collide with other blocks, meaning that this measure of stability is a good test for if a structure has collapsed \cite{abraham2023utilizing}.

After reviewing these metrics, we notice cases in which blocks sustain damage but are not destroyed. Consequently, under the ``Block Destruction" metric, the level is considered stable, while under the ``Block Velocity" metric, it is flagged as unstable. In some instances, blocks have minor movements and are displaced, but the overall shape remains intact. Some of these levels exhibit minimal movement after loading and then stabilize after sustaining minor damage. However, others sustain significant damage, and despite blocks not being destroyed, the shape of the level can change drastically.

To address these scenarios, we introduce a third metric: ``Block Damage."  ``Block Damage" determines that a structure is stable if the total damage sustained by all blocks is less than or equal to zero (the default damage value in the work by Abraham et al. starts at -1). This damage is quantified by the physics engine. Note that the ``Block Destruction" and ``Block Damage" measures are weaker versions of the ``Block Velocity" measure (i.e., any structure classified as stable by ``Block Velocity" is, in the vast majority of cases, also classified as stable by the other two measures).

These three stability metrics provide a multi-faceted view of a level's playability, enabling a more granular analysis of our repair pipeline's effectiveness.

\begin{figure}[ht]
    \centering
    \begin{subfigure}[t]{0.18\textwidth}
        \centering
        \includegraphics[width=\textwidth]{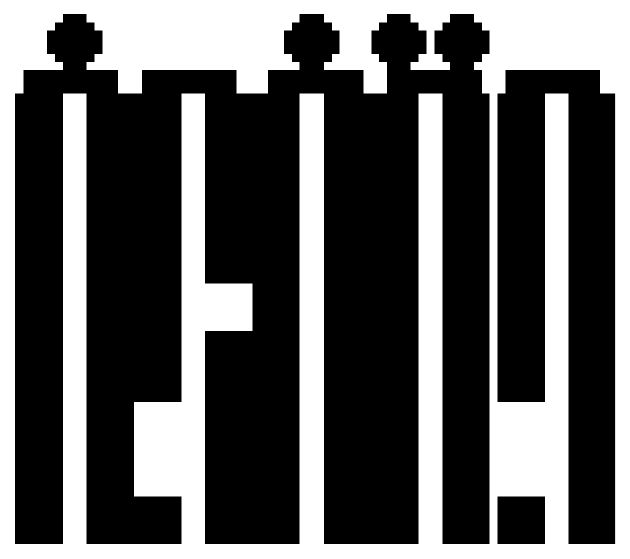}
        \caption{Original Unstable Structure}
        \label{fig:yolo-original}
    \end{subfigure}
    \hspace{0.04\textwidth} 
    \begin{subfigure}[t]{0.18\textwidth}
        \centering
        \includegraphics[width=\textwidth]{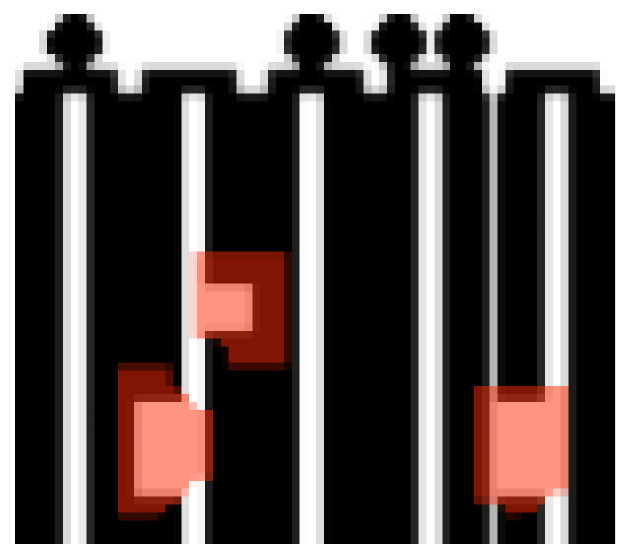}
        \caption{Predicted Gaps by YOLOv8m-Seg}
        \label{fig:yolo-predicted}
    \end{subfigure}
    
    \caption{A visual demonstration of the YOLOv8m-Seg model's gap detection capability. (a) The binary image of an unstable level is given as input. (b) The model outputs a segmentation mask (shown as red overlays) that identifies the location and shape of structural gaps.}
    \label{fig:yolo-example}
\end{figure}

\subsection{Comparing Model Architectures}

To select the optimal backbone for our repair pipeline, we conduct a comparative evaluation of three prominent segmentation architectures: U-Net, SegFormer, and YOLOv8m-Seg. Each model is trained on identical data splits for up to 100 epochs, utilizing an early stopping mechanism to halt training once performance plateaued. The performance of each model is evaluated using precision, recall, F1-Score, and validation loss.

Our results identify YOLOv8m-Seg as the best performing model. It consistently achieve the highest F1-Score (0.824) and the lowest validation loss (1.596), demonstrating a superior balance of accuracy and training stability. SegFormer also shows strong performance, reaching a peak F1-Score of 0.796 with smooth convergence. In contrast, while U-Net obtains the highest recall (0.847), its lower precision results in a less competitive overall F1-Score. The learning progress for each model is visualized in Figure \ref{fig:training-metrics}, and a detailed breakdown of peak performance metrics is provided in Table 2.

\begin{table}[htb]
\centering
\resizebox{1\columnwidth}{!}{%
\begin{tabular}{|l|c|c|c|c|c|}
\hline
\textbf{Model} & \textbf{Epoch} & \textbf{Precision} & \textbf{Recall} & \textbf{F1 Score} & \textbf{Loss} \\
\hline
YOLOv8m     & 33 & 0.837 & 0.811 & 0.824 & 1.596 \\
SegFormer   & 65 & 0.798 & 0.795 & 0.796 & 1.932 \\
UNet        & 54 & 0.690 & 0.847 & 0.760 & 1.721 \\
\hline
\end{tabular}%
}
\caption{Performance comparison of three segmentation models at their best epochs, evaluated using precision, recall, F1 score, and training segmentation loss.}
\label{tab:model-performance}
\end{table}

While SegFormer and UNet remain viable alternatives for applications prioritizing pixel-level fidelity or architectural modularity, we select YOLOv8m-Seg as the backbone for our repair pipeline due to its higher segmentation accuracy and lower loss. To quantify this, we also measure the object detection metrics previously defined, finding that the model achieved a good mAP50 of 0.862 and a mAP50-95 of 0.409. This proven ability to not only segment but also accurately locate gaps, as visually demonstrated in Figure \ref{fig:yolo-example}, confirms it as the most suitable architecture for our task.

\subsection{Efficacy of the Repair Pipeline}

After selecting YOLOv8m-Seg as the optimal backbone, we evaluate the efficacy of the complete repair pipeline. This evaluation was designed to answer two primary questions:

1) How effectively does our method increase the number of stable, playable levels?

2) What are the characteristics and impact of unsuccessful repairs?

To this end, we apply our repair process to the full set of levels generated using the methods of Abraham and Stephenson, which are not part of the training or validation sets.

\subsubsection{Quantitative Repair Success}

To provide a comprehensive view of our pipeline's effectiveness, we evaluate the repair outcomes against each of our three stability metrics, which vary in strictness. This multi-faceted analysis demonstrates how the definition of ``stability" impacts the perceived success rate.
For this analysis, we calculate two key metrics to quantify the performance of our method.

\begin{enumerate}
    \item Repair Rate: This metric measures the direct effectiveness of our pipeline on the pool of unstable levels. It is defined as the proportion of initially unstable levels that were successfully stabilized by our repair process.

    \[
    \textnormal{Repair Rate} =
    \frac{
      \begin{array}{c}
        \textnormal{Number of Levels} \\
        \textnormal{Successfully Repaired}
      \end{array}
    }{
      \begin{array}{c}
        \textnormal{Total Number of } \\
        \textnormal{Initially Unstable Levels}
      \end{array}
    }
    \]

    \item Stability Growth Factor: This metric measures the multiplicative increase in the total yield of usable levels. It is calculated by dividing the final count of stable levels by the initial count, showing how many times larger the set of stable levels became after our repairs.
    \[
    \textnormal{Stability Growth Factor} =
    \frac{
      \begin{array}{c}
        \textnormal{Total Stable Levels} \\
        \textnormal{After Repair}
      \end{array}
    }{
      \begin{array}{c}
        \textnormal{Total Stable Levels} \\
        \textnormal{Before Repair}
      \end{array}
    }
    \]

\end{enumerate}

\begin{table*}[t!]
\centering
\begin{tabular}{|l|r|r|r|r|r|r|}
\hline
\textbf{Metric} & \textbf{Initial Unstable} &  \textbf{Stabilized} & \textbf{Repair Rate} & \textbf{Initial Stable} & \textbf{Final Stable} & \textbf{Growth Factor} \\
\hline
Block Velocity    & 7,055 & 1,254 & 17.8\% & 945  & 2,199 & 2.33 \\
Block Damage      & 6,259 & 1,452 & 23.2\% & 1,741 & 3,193 & 1.83 \\
Block Destruction & 4,533 & 2,051 & 45.3\% & 3,467 & 5,518 & 1.59 \\
\hline
\end{tabular}
\caption{Summary of Repair Outcomes Across Three Stability Metrics.}
\label{tab:repair-results}
\end{table*}

Together, these metrics provide a complete picture of our method's performance, showing both its efficiency in fixing broken content and its overall impact on the total number of playable levels. Our results are summarized in Table 3.

First, we evaluate the results using our most stringent metric, “Block Velocity.” We began with 7,055 levels that were flagged as unstable by this metric. After applying our automated repair process, we re-evaluate this same set of levels. Our results show that the pipeline successfully stabilized 1,254 of them, achieving a Repair Rate of 17.8\%. This result demonstrates a significant and practical improvement, increasing the total count of levels stable under the ``Block Velocity" metric from 945 to 2,199. It represents a substantial improvement over the baseline: out of 8,000 GAN-generated levels, only 945 were initially stable, and our approach increased this to over 2,000. This corresponds to a Stability Growth Factor of 2.33, confirming a significant gain in the total number of playable levels.

Next, we assess the pipeline's performance using our “Block Damage” metric, which captures more nuanced instabilities. This metric initially identifies 6,259 unstable levels. After applying the repair process, 1,452 of these levels are successfully stabilized, achieving a Repair Rate of 23.2\%. This increases the total count of stable levels under this metric from 1,741 to 3,193. This corresponds to a Stability Growth Factor of 1.83, demonstrating a substantial improvement in the number of playable levels.

Finally, we assess the pipeline's performance using the more lenient “Block Destruction” metric. Under this definition, a smaller set of 4,533 levels from the initial pool are considered unstable. When our repair process is applied, 2,051 of these are successfully stabilized, achieving a high Repair Rate of 45.3\%. This increases the total count of stable levels under this metric from 3,467 to 5,518. While this corresponds to a Stability Growth Factor of 1.59, the growth is more modest compared to our other metrics. This is because the initial number of stable levels is already high.

\begin{figure*}[ht]
    \centering
    \begin{subfigure}[t]{0.3\textwidth}
        \centering
        \includegraphics[width=\textwidth]{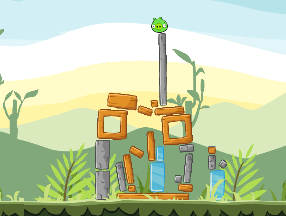}
        \caption{Initial Unstable Level}
        \label{fig:unstable-initial}
    \end{subfigure}
    \hfill
    \begin{subfigure}[t]{0.3\textwidth}
        \centering
        \includegraphics[width=\textwidth]{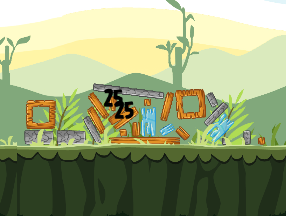}
        \caption{Simulation Result of Unstable Level}
        \label{fig:unstable-result}
    \end{subfigure}
    \hfill
    \begin{subfigure}[t]{0.3\textwidth}
        \centering
        \includegraphics[width=\textwidth]{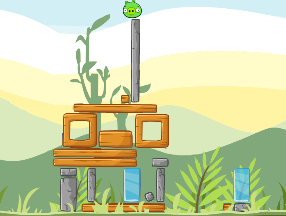}
        \caption{Repaired Stable Level}
        \label{fig:repaired-stable}
    \end{subfigure}
    \caption{An illustration of a successful repair. (a) The initial AI-generated level is flagged as unstable. (b) When the physics simulation runs, the structure collapses. (c) After processing by our system, the level is re-evaluated and confirmed to be stable. Some blocks have changed shape due to inconsistencies in the GAN decoder, as discussed in the 'Limitations and Challenges' section.}
    \label{fig:repair-example}
\end{figure*}

These varying rates offer a nuanced perspective on our method's performance. The modest repair rates for the stricter ``Block Velocity" and ``Block Damage" metrics underscore the difficulty of the task and the limitations of our model. However, the high Stability Growth Factor in these same cases is still noteworthy. This suggests that while a successful repair is not the most frequent outcome for these challenging instabilities, each successful fix provides a valuable contribution to the final dataset, especially when the initial number of stable levels is low.
Figure \ref{fig:repair-example} illustrates a successful application of our repair pipeline. It shows a level that was initially flagged as unstable according to our metrics and then repaired by our system.

We believe our model's varying success rates across the different metrics can be explained by the nature of the instabilities each metric captures. The stricter ``Block Velocity" and ``Block Damage" metrics are often triggered by slight movements caused by small, well-defined gaps in the generated levels. Our model performs well in these cases, as it is specifically designed to identify and fill such gaps. However, when blocks are destroyed (triggering the ``Block Destruction" metric), it often signals a major structural collapse. These catastrophic failures usually cannot be fixed by simply filling gaps and likely require more substantial modifications beyond the scope of our current method.

As noted earlier in our methodology, the risk of false positives, which are instances where a repair process might negatively affect an already stable level, is inherently mitigated by the design of our two-stage pipeline. The initial stability assessment acts as a gatekeeper, ensuring that only levels flagged as unstable are passed to the repair model. Consequently, the repair function is not performed on stable levels, which focuses the process exclusively on improving unplayable content without the danger of corrupting already-valid levels.

\subsubsection{Analysis of Unsuccessful Repairs}

To better understand the limitations of our method, we analyze the set of levels that remained unstable after a repair attempt. The failures can be broadly categorized into two primary modes.

In approximately 24\% of unsuccessful repairs, the model fails to confidently identify any structural gaps above its confidence threshold, even though the level is unstable. This suggests that a significant portion of instabilities are not caused by detectable gaps but by more complex architectural flaws (e.g., poor weight distribution) that our segmentation-based approach is not designed to address. This observation is consistent with the validation results of the model and highlights a clear area for future work.

In the remaining cases, the model correctly identifies and fills one or more gaps, but this single intervention is insufficient to stabilize the entire structure. This often occurs in levels with multiple points of weakness, suggesting that some instabilities require more complex solutions than just filling gaps, such as modifying existing blocks or performing multi-step repairs.
We also compare the average damage of the failed repairs against the original unrepaired levels to measure the impact of our model even in these unsuccessful cases. Our analysis reveals that even when a repair does not achieve full stability, it often makes the levels more stable than they were originally. On average, these levels demonstrate a 25.1\% reduction according to the Block Damage metric (from 51.99 to 38.95). Similarly, the average count of destroyed blocks, which informs our Block Destruction metric, decreases by 11.5\% (from 5.2 to 4.6). This indicates that even unsuccessful repairs often reduced the overall severity of the structural collapse, suggesting our method can provide value even when it does not fully solve the instability.

In summary, these findings confirm that our repair process effectively reduces structural weaknesses in AI-generated levels. While the pipeline could not repair every unstable instance, its success and almost doubling the number of playable levels demonstrates its practical value. Moreover, the discovery that unsuccessful repairs still yield a partial, mitigating effect by reducing average block damage highlights the robustness of our method. These outcomes validate the potential of our segmentation-based approach for enhancing the quality and reliability of procedurally generated levels.

\section{Challenges and Limitations}

While our segmentation-based repair pipeline demonstrates success in improving the stability of AI-generated Angry Birds levels, it is essential to acknowledge its limitations and the challenges encountered. These aspects not only define the boundaries of our current method but also highlight promising directions for future research.

A primary limitation stems from the design of our training dataset. To create a supervised learning task with a clear ground truth, we generated gaps by removing blocks from otherwise stable structures. However, this approach does not fully represent the more chaotic and complex failure modes of procedurally generated content. Automatically generated instabilities can arise from more than just clean gaps. They may result from poor weight distribution, awkward object intersections, or a series of minor misalignments that collectively compromise the structure. As a result, our model is specialized in identifying and fixing gaps, which, as our analysis shows, are not the sole cause of instability.

Another challenge lies in the limitations of image-based segmentation. Our model identifies gaps based purely on the visual and spatial layout of level components, with no explicit understanding of underlying physics, material properties (such as weight, friction, or durability), or object-to-object forces. This explains the ``gap detection failure” scenario, where the model found no gap to fix because the instability was not caused by a visually apparent hole. Additionally, our current repair process defaults to filling gaps with wood. While our findings suggest that changing the material does not significantly affect stability, this uniform approach can reduce material diversity in repaired levels. To improve this, a material selection algorithm could be introduced.

Our work operates as a post-processing step and thus inherits complexities from the tools it relies on. A significant challenge is the non-determinism in the GAN decoder that our work used. As shown in Figure \ref{fig:repair-example}, the decoder does not always reproduce the same structure for a given XML level file; for example, it may replace a single large block with two smaller ones. However, we examined whether repeated renderings of the same XML input could lead to inconsistent stability outcomes due to this non-determinism. While the decoder occasionally introduces small visual differences (e.g., splitting one block into two), our experiments showed that these variations do not affect the stability classification. Specifically, if a level is unstable, it remains unstable across repeated renderings; likewise, if a repaired level is stabilized, it consistently remains stable even with minor shape differences. This confirms that our stability evaluations and repairs are robust to such decoding variability. Nevertheless, this variability introduces an additional source of error and inconsistency into the repair pipeline. To mitigate this, future work could focus on standardizing the decoding process.

\section{Conclusions and Future Work}

Ensuring the quality and playability of procedurally generated content remains a critical bottleneck, particularly for physics-based games where structural stability is essential. In this paper, we address this challenge by introducing an automated, post-processing repair pipeline designed to salvage unstable levels generated by AI. By training an object segmentation model, YOLOv8m-Seg, to identify and fill structural gaps, our method successfully increases the yield of playable Angry Birds levels from an existing GAN-based generator.

Despite these successes, our work also highlights several key challenges that pave the way for future research. Our current model is trained to fix instabilities by filling visually apparent gaps. Future work should aim to address more complex failure modes by moving beyond a purely visual approach. This could involve developing hybrid models to diagnose instabilities caused by poor weight distribution or material properties, and not just missing blocks. Future research could also explore semi-supervised or self-supervised learning techniques to train models on a wider variety of instabilities without requiring manually crafted ``gap" examples. We also observed inconsistencies in decoding levels. Since the overall shape of the structures remained largely unchanged, we do not expect these inconsistencies to have had a significant effect. Nonetheless, addressing this issue could lead to more accurate results, making it a promising area for improvement in future work.

While this work focused on Angry Birds, we believe the core concept of a segmentation-based repair agent operating on image inputs and 2D representations can be adapted to other games with visual structures. For example, this framework could be extended to 2D platformers such as Super Mario Bros. to detect and fix issues like impassable jumps, broken paths, or unreachable areas. Testing the generalizability of this approach across different game mechanics and environments remains a vital next step.

\section{Acknowledgements}
This research was supported by the Natural Sciences and Engineering Research Council of Canada (NSERC) Discovery Grant. We thank members of the Serious Games Research Group and the anonymous reviewers for their feedback. We thank Hosein Beheshtifard for contributing to an earlier concept of this research.

\bibliography{aaai25}

\end{document}